\RequirePackage{fix-cm}
\documentclass[smallextended]{svjour3}       

\usepackage[hyphens]{url}  
\usepackage{subfig}
\usepackage{caption} 
\usepackage[misc]{ifsym}

\smartqed  
\usepackage{graphicx}
\usepackage{amsmath,amsfonts,amssymb}
\usepackage[linkcolor=blue,colorlinks=true]{hyperref} 
\usepackage{bm}
\usepackage{algorithm}
\usepackage{algorithmic} 
\usepackage{color}
\usepackage{booktabs}
\usepackage{diagbox}
\usepackage{multirow} 
\usepackage{mathptmx}      
\usepackage{float}
\usepackage{makecell}  

\newcommand{\bX}{\mathbf{X}}
\newcommand{\bx}{\mathbf{x}}

\newcommand{\bY}{\mathbf{Y}}
\newcommand{\bU}{\mathbf{U}}
\newcommand{\bS}{\mathbf{S}}

\newcommand{\citep}[1]{\cite{#1}}

\begin{document}

\title{Fast Estimation and Model Selection for Penalized PCA via Implicit Regularization of SGD}
\title{AgFlow: Fast Model Selection of Penalized PCA via Implicit Regularization Effects of Gradient Flow}



\author{Haiyan Jiang\and Haoyi Xiong~\Letter\and Dongrui Wu\and Ji Liu\and Dejing Dou
}

\authorrunning{Haiyan Jiang, Haoyi Xiong, Dongrui Wu, Ji Liu, and Dejing Dou} 

\institute{H. Jiang\and H. Xiong~\Letter\and J. Liu\and D. Dou \at
              Big Data Lab, Baidu Research \\
              \email{xionghaoyi@baidu.com}           
           \and
           D. Wu \at
           School of Artificial Intelligence and Automation\\
             Huazhong University of Science and Technology
}

\date{Received: date / Accepted: date}

\maketitle

\begin{abstract}


Principal component analysis (PCA) has been widely used as an effective technique for feature extraction and dimension reduction.
In the High Dimension Low Sample Size (HDLSS) setting, one may prefer modified principal components, with penalized loadings, and automated penalty selection by implementing model selection among these different models with varying penalties.
The earlier work~\citep{zou2006sparse,gaynanova2017penalized} has proposed penalized PCA,
indicating the feasibility of model selection in $\ell_2$-penalized PCA through the solution path of Ridge regression, however, it is extremely time-consuming because of the intensive calculation of matrix inverse.
In this paper, we propose a fast model selection method for penalized PCA, named Approximated Gradient Flow (\texttt{AgFlow}), which lowers the computation complexity through incorporating the implicit regularization effect introduced by (stochastic) gradient flow~\citep{ali2019continuous,ali2020implicit} and obtains the complete solution path of $\ell_2$-penalized PCA under varying $\ell_2$-regularization. 
We perform extensive experiments on real-world datasets. 
\texttt{AgFlow} outperforms existing methods (Oja~\citep{oja1985stochastic}, Power~\citep{hardt2014noisy}, and Shamir~\citep{shamir2015stochastic} and the vanilla Ridge estimators) in terms of computation costs.

\end{abstract}

\section{Introduction} \label{intro}
Principal component analysis (PCA)~\citep{jolliffe1986principal,dutta2019nonconvex} is widely used as an effective technique for feature transformation, data processing and dimension reduction in unsupervised data analysis, with numerous applications in machine learning and statistics such as handwritten digits classification~\citep{lecun1995comparison,hastie2009elements}, human faces recognition~\citep{huang2008labeled,mohammed2011human}, and gene expression data analysis~\citep{yeung2001principal,zhu2007markov}. Generally, given a data matrix $\bX \in \mathbb{R}^{n\times d}$ , where $n$ refers to the number of samples and $d$ refers to the number of variables in each sample, PCA can be formulated as a problem of projecting samples to a lower $d'$-dimensional subspace ($d'\ll d$) with variances maximized.
%
%
To achieve the goal, numerous algorithms, such as Oja's algorithm~\citep{oja1985stochastic}, power iteration algorithm~\citep{hardt2014noisy}, and stochastic/incremental algorithm~\citep{shamir2015stochastic,arora2012stochastic,mitliagkas2013memory,de2015global} have been proposed, and the convergence behaviors of these algorithms have also been intensively investigated. In summary, given the matrix of raw data samples, the eigensolvers above output $d'$-dimensional vectors which are linear combinations of the original predictors, projecting original samples into the $d'$-dimensional subspace desired while capturing maximal variances.

\begin{figure*}[h!]
\centering
\subfloat[Validation Accuracy with $\ell_2$-penalized PCA ]{\includegraphics[width=0.45\textwidth]{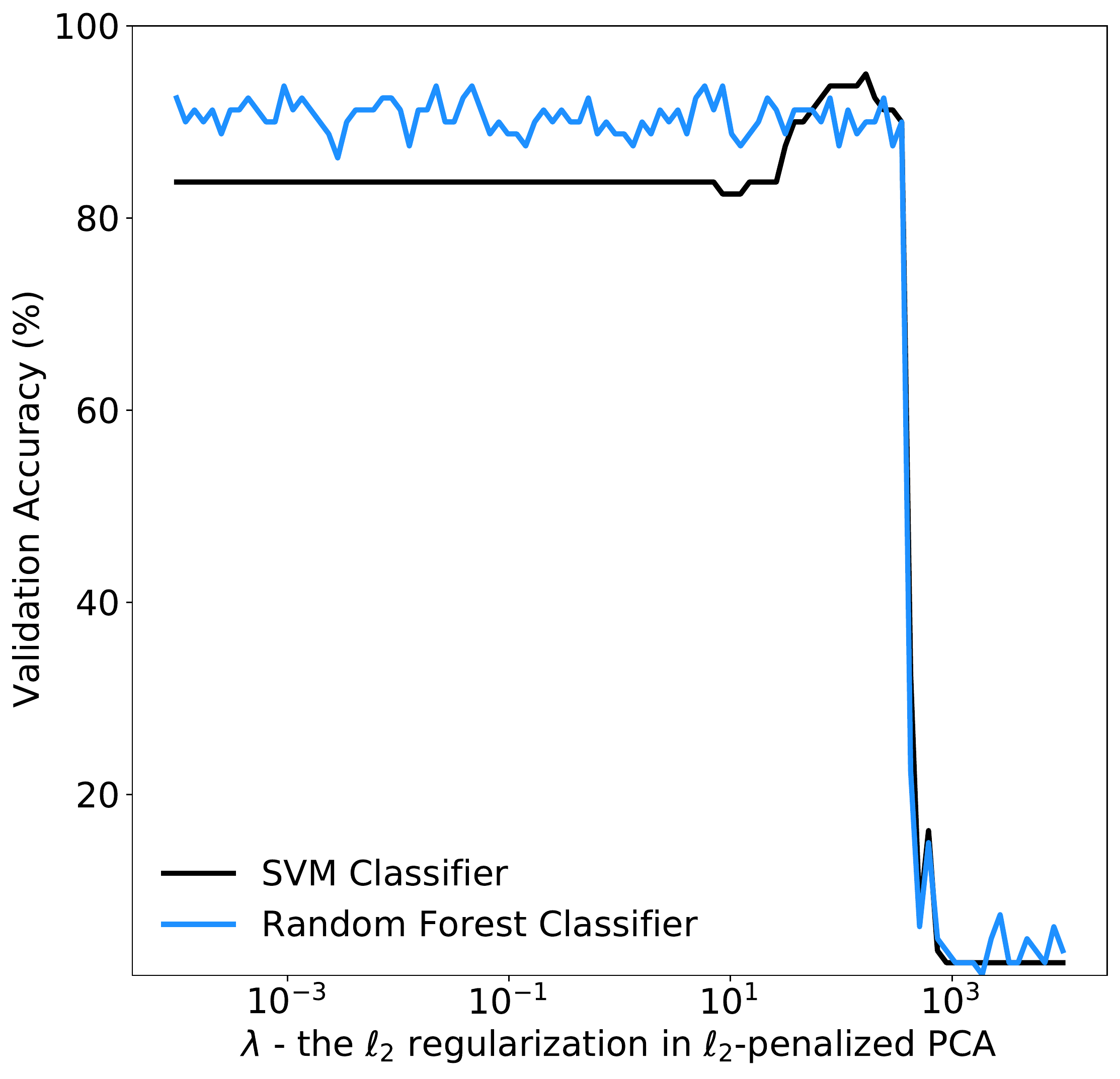}}\
\subfloat[Testing Accuracy with $\ell_2$-penalized PCA ]{\includegraphics[width=0.45\textwidth]{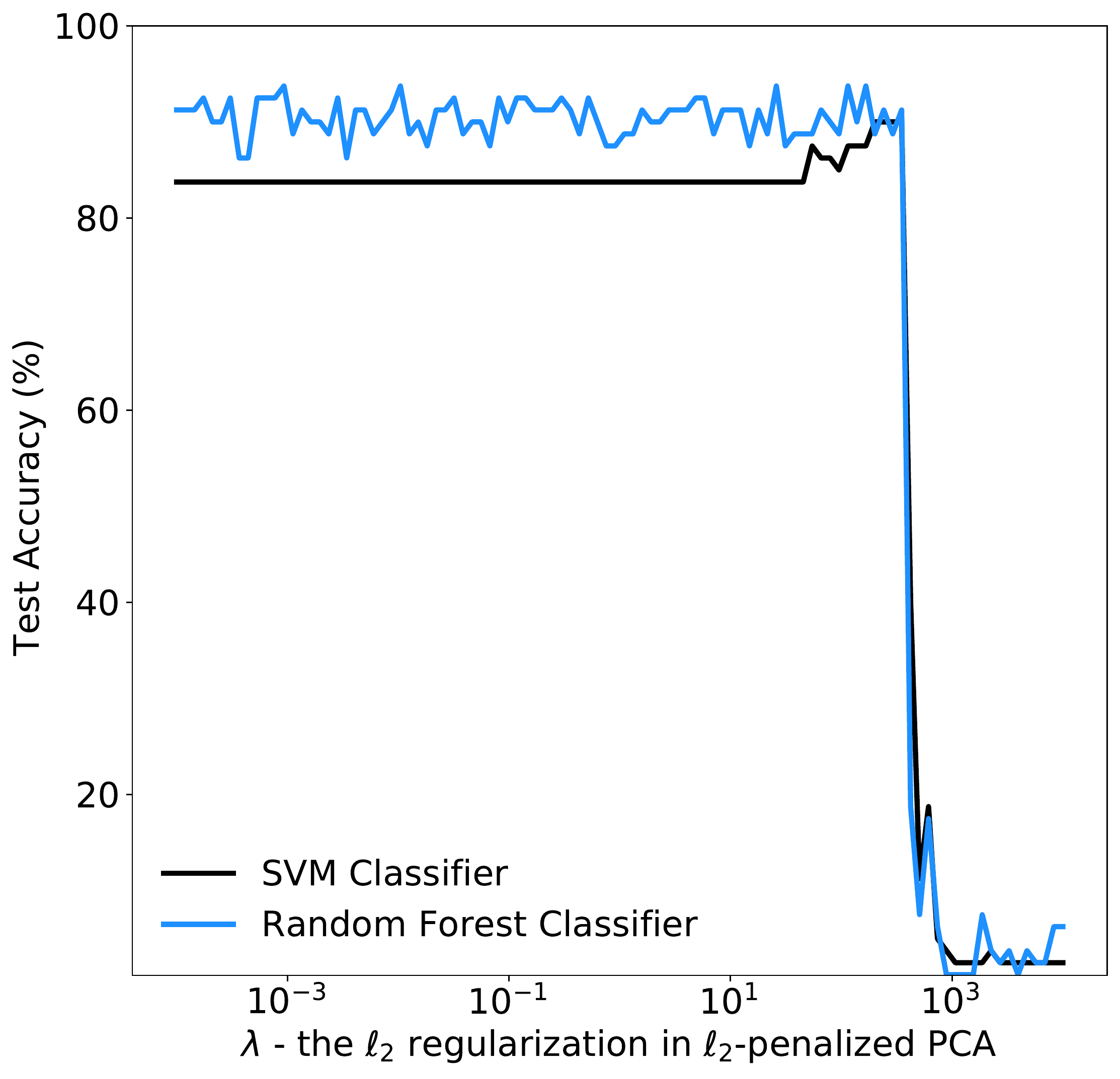}}
\caption{A Running Example of Performance Tuning for Classification with $\ell_2$-penalized PCA Dimension-Reduced Data. We reduce the dimension of FACES dataset~\cite{huang2008labeled} from $d=4096$ to $d'=20$, using $\ell_2$-penalized PCA (Ridge-based), and try to select models over the solution path of varying $\lambda$ for classification tasks. The model (i.e., the $\ell_2$-penalized PCA estimation with a fixed $\lambda$) that achieves better performance in validation set empirically works better on the testing set.}
\label{fig:ridge-toy}
\end{figure*}

In addition to the above estimates of PCA, penalized PCA has been proposed~\citep{zou2006sparse,gaynanova2017penalized,witten2009penalized,lee2012principal} to improve its performance using regularization. For example,~\cite{zou2006sparse} introduced a direct estimation of $\ell_2$-penalized PCA using Ridge estimator (see also in Theorem.~1 in Section 3.1 of~\citep{zou2006sparse}), where an $\ell_2$-regularization hyper-parameter (denoted as $\lambda$) has been used to balance the error term for fitting and the penalty term for regularization. Whereas an $\ell_1$-regularization is usually introduced for achieving sparsity~\cite{gaynanova2017penalized}. Though the effects of $\lambda$ in $\ell_2$-penalized PCA would be waived by normalization when the sample covariance matrix is non-singular (i.e., $d<n$), the penalty term indeed regularizes the sample covariance matrix~\citep{witten2009covariance} for a stable inverse under High Dimension Low Sample Size (HDLSS) settings. 
Therefore, it is desirably necessary to do model selection to get the optimal $\lambda$ on the solution path, where the solution path is formed by all the solutions corresponding to all the candidate $\lambda$-s in the $\ell_2$-penalized problem, and each model is determined by the parameter $\lambda$.
Thus, given the datasets for training and validation, it is only needed to retrieve the complete solution path~\cite{friedman2010regularization,zou2005regularization} for penalized PCA using the training dataset, where each solution corresponds to an Ridge estimator, and iterate every model on the solution path for validation and model selection. An example of $\ell_2$-penalized PCA for dimension reduction over the solution path is listed in Fig.~\ref{fig:ridge-toy}, where we can see that both the validation and testing accuracy are heavily affected by the value of $
\lambda$, the $\ell_2$ regularization in $\ell_2$-penalized PCA, no matter what kind of classifier is employed after dimension reduction.

While a solution path for penalized PCA is highly required, the computational complexity of estimating a large number of models using 
grid searching 
of hyper-parameters is usually unacceptable. Specifically, to obtain the complete solution path or the models for $\ell_2$-penalized PCA, it is needed to repeatedly solve the Ridge estimator with a wide range of values for $\lambda$, where the matrix inverse to get a shrunken sample covariance matrix is required in $O(d^3)$ complexity for every possible setting of $\lambda$. 
To lower the complexity, inspired by the recent progress on implicit regularization effects of gradient descent (GD) and stochastic gradient descent (SGD) in solving Ordinary Least-Square (OLS) problems~\citep{ali2019continuous,ali2020implicit}, we propose a fast model selection method, named Approximated Gradient Flow (\texttt{AgFlow}), which is an efficient and effective algorithm to accelerate model selection of $\ell_2$-penalized PCA with varying penalties.

\paragraph{\textbf{Our Contributions. }} We make three technical contributions as follows.
\begin{itemize}
\item We study the problem of lowering the computational complexity while accelerating the model selection for penalized PCA under varying penalties, where we particularly pay attention to $\ell_2$-penalized PCA via the commonly-used Ridge estimator~\cite{zou2006sparse} under High Dimension Low Sample Size (HDLSS) settings. 

\item We propose \texttt{AgFlow} to do fast model selection in $\ell_2$-penalized PCA with $O(K\cdot d^2)$ complexity, where $K$ refers to the total number of models estimated for selection, and $d$ is the number of dimension. More specifically, \texttt{AgFlow} first adopts algorithms in~\cite{shamir2015stochastic} to sketch $d'$ principal subspaces, then retrieves the complete solution path of the corresponding loadings for every principal subspace. Especially, \texttt{AgFlow} incorporates the implicit $\ell_2$-regularization of approximated (stochastic) gradient flow over the Ordinary Least Squares (OLS) to screen and validate the $\ell_2$-penalized loadings, under varying $\lambda$ from $+\infty\to 0^+$, using the training and validation datasets respectively.

\item We conduct extensive experiments to evaluate \texttt{AgFlow}, where we compare the proposed algorithm with vanilla PCA, including~\citep{hardt2014noisy,shamir2015stochastic,oja1985stochastic} and $\ell_2$-penalized PCA via Ridge estimator~\citep{zou2006sparse} on real-world datasets. Specifically, the experiments are all based on HDLSS settings, where a limited number of high-dimensional samples have been given for PCA estimation and model selection. The results showed that the proposed algorithm can significantly outperform the vanilla PCA algorithms~\citep{hardt2014noisy,shamir2015stochastic,oja1985stochastic} with better performance on validation/testing datasets gained by the flexibility of performance tuning (i.e., model selection).
On the other hand,~\texttt{AgFlow} consumes even less computation time to select models from 50 times more models compared to Ridge-based estimator~\citep{zou2006sparse}.
\end{itemize}
Note that we don't intend to propose the ``off-the-shelf'' estimators to reduce the computational complexity of PCA estimation. Instead, we study the problem of model selection for $\ell_2$-regularized PCA, where we combine the existing algorithms~\cite{zou2006sparse,ali2019continuous,shamir2015stochastic} to lower the complexity of model selection and accelerate the procedure. The unique contribution made here is to incorporate with the novel continuous-time dynamics of gradient descent (gradient flow)~\cite{ali2019continuous,ali2020implicit} to obtain the time-varying implicit regularization effects of $\ell_2$-type for PCA model selection purposes. 



\paragraph{\textbf{Notations. }}
The following key notations are used in the rest of the paper.
Let $\mathbf{x} \in \mathbb{R}^{d}$ be the $d$-dimensional predictors and $y \in \mathbb{R}$ be the response, and denote $\bX =[\mathbf{x}_1, \cdots, \mathbf{x}_n]^\top = [X_1, \cdots, X_d]$ and $Y = [y_1, \cdots, y_n]^\top$, where $n$ is the sample size and $d$ is the number of variables. Without loss of generality, assume $X_{j},\ j=1, \cdots, d$ and $Y$ are centered. Given a $d$-dimensional vector $\mathbf{z} \in \mathbb{R}^{d}$, denote the $\ell_2$ vector-norm $ \|\mathbf{z}\|_2 = \left( \sum_{i=1}^d |z_i|^2 \right)^{1/2} $.

\section{Preliminaries}\label{sec:prelim}

In this section, firstly the Ordinary Least Squares and Ridge Regression is briefly introduced, then followed with the formulation that PCA is rewritten as a regression-type optimization problem with an explicit $\ell_2$ regularization parameter $\lambda$, and lastly ended up with the introduction of the implicit regularization effect introduced by the (stochastic) gradient flow.

\subsection{Ordinary Least Squares and Ridge Regression}

Let $\bX \in \mathbb{R}^{n \times d}$ and $Y \in \mathbb{R}^{n}$ be a matrix of predictors (or features) and a response vector, respectively, with $n$ observations and $d$ predictors. Assume the columns of $\bX$ and $Y$ are centered.
Consider the ordinary least squares (linear) regression problem
\begin{equation}\label{eq:ols} 
   \hat\beta_\mathrm{OLS}=\mathop{\arg\min}_{\beta \in \mathbb{R}^{d}}\
    \frac{1}{2n} \| Y - \bX \beta \|_2^2 \ .
\end{equation}
To enhance the solution of OLS for linear regression, regularization is commonly used as a popular technique in optimization problems in order to achieve a sparse solution or alleviate the multicollinearity problem~\citep{friedman2010regularization,zou2005regularization,tibshirani1996regression,fan2001variable,yuan2006model,candes2007dantzig}. 
Recently an enormous amount of literature has focused on the related regularization methods, such as the lasso~\citep{tibshirani1996regression} which is friendly to interpretability with a sparse solution, 
the grouped lasso~\citep{yuan2006model} where variables are included or excluded in groups, the elastic net~\citep{zou2005regularization} for correlated variables which compromises $\ell_1$ and $ \ell_2$ penalties, the Dantzig selector~\citep{candes2007dantzig} which serves as a slightly modified version of the lasso, and some variants~\citep{fan2001variable}.

The ridge regression is the $\ell_2$-regularized version of the linear regression in Eq.~(\ref{eq:ols}), imposing an explicit $\ell_2$ regularization on the coefficients~\citep{hoerl1970ridge,hoerl1975ridge}.
Thus, the ridge estimator $\hat{\beta}_{\text{ridge}}(\lambda)$, a penalized least squares estimator, can be obtained by minimizing the ridge criterion
\begin{equation}\label{eq:ridge-reg}
    \hat{\beta}_{\text{ridge}}(\lambda)
    = \mathop{\arg \min}_{\beta  \in \mathbb{R}^d} 
    \left\{ \frac{1}{2n} \|Y - \bX \beta \|_2^2 
    + \frac{\lambda}{2}\| \beta\|_2^2 \right\}.
\end{equation}
The solution of the ridge regression has an explicit closed-form, 
\begin{equation}\label{eq:ridge-closedform}
    \hat{\beta}_{\text{ridge}}(\lambda)
    = (\bX^{\top} \bX + n \lambda \mathbf{I})^{-1} \bX^{\top} Y.
\end{equation}
We can see that the ridge estimator, Eq.~(\ref{eq:ridge-closedform}), applies a type of shrinkage in comparison to the OLS solution 
$\hat{\beta}_{\text{OLS}} = (\bX^{\top} \bX)^{-1} \bX^{\top}Y$, 
which shrinks the coefficients of correlated predictors towards each other and thus alleviates the multicolinearity problem.

\subsection{PCA as Ridge Regression}
PCA can be formulated as a regression-type optimization problem which was first proposed by~\citep{zou2006sparse}, where the loadings could be recovered by regressing the principal components on the $d$ variables given the principal subspace.

Consider the $j^{th}$ principal component. Let $\bY^j$ be a given $n \times 1$ vector referring to the estimate of the $j^{th}$ principal subspace. For any $\lambda \geq 0$, the Ridge-based estimator (Theorem.~1 in~\cite{zou2006sparse}) of $\ell_2$-penalized PCA is defined as
\begin{equation}\label{eq:pca-ridge}
    \bar{\beta}^j(\lambda) = \mathop{\arg\min}_{\beta \in \mathbb{R}^d} 
    \left\{\frac{1}{2n} \left\| \bY^j - \bX \beta\right\|_2^2 
    + \frac{\lambda}{2} \|\beta\|_{2}^2 \right\} . 
\end{equation}
Obviously, the estimator above highly depends on the estimate of the principal subspace $\bY^{j}$. Given the original data matrix $\bX=[\mathbf{x}_1,\cdots,\mathbf{x}_n]^\top$, we could obtain its singular value decomposition as $\bX=\mathbf{USV}^\top$ and the estimate of subspace could be $\bY^j=\bU_j\bS_j$, where $\bU_j$ and $\bS_j$ refers to the $j^{th}$ columns of the corresponding matrices, respectively. Then the normalized vector can be used as the penalized loadings of the $j^{th}$ principal component
\begin{equation}\label{normalize}
    \hat\beta^{j}({\lambda})=\frac{\bar\beta^j(\lambda)}{\|\bar\beta^j(\lambda)\|_2} .
\end{equation}
Note that, when the sample covariance matrix $\frac{1}{n}\bX^\top\bX$ is nonsingular ($d\leq n$), $\hat\beta^j(\lambda)$ would be invariant on $\lambda$ and $\hat\beta^j(\lambda)\propto\mathbf{V}_j$. When the sample covariance matrix is singular ($d>n$), the $\ell_2$-norm penalty would regularize the inverse of shrunken covariance matrix~\citep{witten2009covariance} with respect to the strength of $\lambda$. 


\subsection{Implicit Regularization with (Stochastic) Gradient Flow. }


The implicit regularization effect of an estimation method means that the method produces an estimate exhibiting a kind of regularization, even though the method does not employ an explicit regularizer~\citep{ali2019continuous,ali2020implicit,friedman2003gradient,friedman2004gradient}.
Consider gradient descent applied to Eq.~(\ref{eq:ols}), with initialization value $\beta_0 = \mathbf{0}$, and a constant step size $\eta > 0$, which gives the iterations
\begin{equation}\label{eq:gd-k}
    \beta_{k} = \beta_{k-1} + \frac{\eta}{n} \bX^{\top} (Y - \bX \beta_{k-1}) \ ,
\end{equation}
for $k=1, 2, 3, \cdots $. 
With simply rearrangement, we get 
\begin{equation}\label{eq:gd-k-2}
  \frac{\beta_{k} - \beta_{k-1}}{\eta}
  = \frac{\bX^{\top} (Y - \bX \beta_{k-1})}{n} \ .
\end{equation}
To adopt a continuous-time (gradient flow) view,
consider infinitesimal step size in gradient descent, i.e., $\eta \to 0 $. The \emph{gradient flow} differential equation for the OLS problem can be obtained with the following equation,
\begin{equation}\label{eq:gd-t}
  \dot{\beta}(t) = \frac{1}{n} \bX^{\top} \left(Y - \bX \beta(t-1)\right),
\end{equation}
which is a continuous-time ordinary differential equation over time $t \geq 0$ with an initial condition $\beta(0) = \mathbf{0}$.
We can see that by setting $\beta(t) = \beta_{k}$ at time $t=k \eta$, the left-hand side of Eq.~(\ref{eq:gd-k-2}) could be viewed as the discrete derivative of $\beta(t)$ at time $t$, which approaches its continuous-time derivative as $\eta \to 0 $.
To make it clear, $\beta(t)$ denotes the continuous-time view, and $\beta_{k}$ the discrete-time view.
\begin{lemma}
With fixed predictor matrix $\bX$ and fixed response vector $Y$, the gradient flow problem in Eq.~(\ref{eq:gd-t}), subject to $\beta(0) = \mathbf{0}$, admits the following exact solution~\citep{ali2019continuous}
\begin{equation}\label{eq:gf-sol}
  \hat{\beta}_\mathrm{gf}(t) = (\bX^{\top} \bX)^{+}
  (\mathbf{I} - \exp(-t \bX^{\top} \bX / n)) \bX^{\top} Y ,
\end{equation}
for all $t \geq 0$. Here $A^{+}$ is the Moore-Penrose generalized inverse of a matrix $A$, and $\exp(A) = I + A + A^2/2! + A^3/3! + \cdots $ is the matrix exponential.
\end{lemma}

In continuous-time, $\ell_2$-regularization corresponds to taking the estimator $\hat{\beta}_\mathrm{gf}(t)$ in Eq.~(\ref{eq:gf-sol}) for any finite value of $t \geq 0$, where smaller $t$ corresponds to greater regularization. 
Specifically, the time $t$ of gradient flow and the tuning parameter $\lambda$ of ridge regression are related by $\lambda = 1/t$. 



\section{The Proposed \texttt{AgFlow} Algorithm}\label{sec:agflow}
In this section, we first formulate the research problem, then present the design of proposed algorithm with a brief algorithm analysis. 


\subsection{Problem Definition for $\ell_2$-Penalized PCA Model Selection} 
We formulate the model selection problem as selecting the empirically-best $\ell_2$-Penalized PCA for the given dataset with respect to a performance evaluator.
\begin{itemize}
    \item $\lambda\in\Lambda\subseteq \mathbb{R}^+$ -- the tuning parameters and the set of possible tuning parameters (which is a subset of positive reals);
    \item $\bX_{\text{train}} \in \mathbb{R}^{n_{\text{train}} \times d}$ and  $\bX_{\text{val}} \in \mathbb{R}^{n_{\text{val}} \times d}$ -- the training data matrix and the validation data matrix, with $n_{\text{train}}$ samples and $n_{\text{val}}$ samples respectively;
    \item $\bY^j$ -- a given $n \times 1$ vector referring to the estimate of the $j^{th}$ principal subspace;
    \item $\hat{\beta}^{j}(\lambda)$ -- the $j^{th}$ projection vector, or the corresponding loading vector of the $j^{th}$ PC, 
    $\hat{\beta}^{j}(\lambda) = \left(\bX^{\top} \bX + n\lambda \mathbf{I}\right)^{-1} \bX^{\top} \bY^{j}$ 
    solution of Eq.~(\ref{eq:pca-ridge}); 
    \item $\hat{\boldsymbol{\beta}}(\lambda) = [\hat{\beta}^{1}(\lambda), \hat{\beta}^{2}(\lambda), \cdots, \hat{\beta}^{d'}(\lambda)] \in \mathbb{R}^{d \times d'}$ -- the projection matrix based on $\ell_2$-penalized PCA with the tuning parameter $\lambda$, where each column $\hat{\beta}^{j}(\lambda)$ is the corresponding loadings of the $j^{th}$ principal component;
    \item $\bX_{\text{train}} \hat{\boldsymbol{\beta}}(\lambda)$ and $\bX_{\text{val}} \hat{\boldsymbol{\beta}}(\lambda)$  -- the dimension-reduced training data matrix and the dimension-reduced validation data matrix, respectively;
    \item $\mathtt{model(\cdot)}:\mathbb{R}^{n_\mathrm{train}\times d'}\to\mathcal{H}$ -- the target learner for performance tuning that outputs a model $h\in\mathcal{H}$ using the dimension-reduced training data $\bX_{\text{train}} \hat{\boldsymbol{\beta}}(\lambda)$. 
    \item $\mathtt{evaluator(\cdot)}:\mathbb{R}^{n_\mathrm{val}\times d'}\times\mathcal{H}\to\mathbb{R}$ -- the evaluator that outputs the reward (a real scalar) of the input model $h$ based on the dimension-reduced validation data $\bX_{\text{val}} \hat{\boldsymbol{\beta}}(\lambda)$.
\end{itemize}
Then the model selection problem can be defined as follows.
\begin{equation}\label{eq:prob-def}
\begin{aligned}
  &\underset{\lambda\in\Lambda}{\text{maximize} } &&\ \mathtt{evaluator}\left(\bX_{\text{val}} \hat{\boldsymbol{\beta}}(\lambda), \ h(\lambda) \right)\ ,  \\
   &\text{ subject to }&&\ h(\lambda) = \mathtt{model}\left(\bX_{\text{train}} \hat{\boldsymbol{\beta}}(\lambda) \right) .
\end{aligned}
\end{equation}
Where
\begin{align}
  & \bar{\beta}^j(\lambda)=\mathop{\arg\min}_{\beta \in \mathbb{R}^d} 
  \left\{ \frac{1}{2n} \left\| \bY^j - \bX \beta\right\|_2^2 + \frac{\lambda}{2} \|\beta\|_{2}^2\right\},\ \text{ for } j=1, \cdots, d', \nonumber \\
  & \hat{\beta}^{j}(\lambda)=\frac{\bar{\beta}^j(\lambda)}{\|\bar{\beta}^j(\lambda)\|_2} .
\end{align}

Note that $\mathtt{model}(\cdot)$ can be any arbitrary target learner in the learning task and $\mathtt{evaluator}(\cdot)$ can be any evaluation function of validation metrics. To make it clear, we take a classification problem as an example, thus the target learner $\mathtt{model}(\cdot)$ can be the support vector machine (SVM) or random forest (RF), and the evaluation function $\mathtt{evaluator}(\cdot)$ can be the classification error.
To solve the above problem for arbitrary learning tasks $\mathtt{model}(\cdot)$ under various validation metrics $\mathtt{evaluator}(\cdot)$, there are at least two technical challenges needing to be addressed,
\begin{enumerate}
    \item \emph{Complexity - } For any given and fixed $\lambda$, the time complexity to solve the $\ell_2$-penalized PCA (for dimension reduction to $d'$) based on the Ridge-regression is $O(d'\cdot d^3)$, as it requests to solve the Ridge regression (to get the Ridge estimator) $d'$ rounds to obtain the corresponding loadings of the top-$d'$ principal components and the complexity to calculate the Ridge estimator in one round is $O(d^3)$.
    \item \emph{Size of $\Lambda$ - } The performance of the model selection relies on the evaluation of models over a wide range of $\lambda$, while the overall complexity to solve the problem should be $O(|\Lambda|\cdot d'\cdot d^3)$. Thus, we need to obtain a well-sampled set of tuning parameters $\Lambda$ that can balance the cost and the quality of model selection.
\end{enumerate}


\begin{algorithm}[h]
\begin{algorithmic}
    \algsetup{linenosize=\normalsize}
    \STATE \textbf{Input data:} 
        \STATE \quad Training data matrix $\bX_{\text{train}} = [\mathbf{x}_1, \cdots ,\mathbf{x}_{n_{\text{train}}}]^\top$
        \STATE \quad Validation data matrix $\bX_{\text{val}} = [\mathbf{x}_1, \cdots ,\mathbf{x}_{n_{\text{val}}}]^\top$
    \STATE \textbf{Parameters:} 
        \STATE \quad Iterations $K$, step size $\eta$, batch size $m$, the reduced dimension $d'$.
    \medskip
    \STATE {\textcolor{blue}{\%Approximate $\hat{\boldsymbol{\beta}}(\lambda)$ with $\hat{\boldsymbol{\beta}}_k$ using Approximated Gradient Flow \%}}
    \FOR{$j = 1, \cdots, d'$}
        \STATE \textbf{Get the $j$th principal subspace:} $\bY^j \gets \mathtt{QuasiPS(\mathbf{X},j)}$.      
        \STATE $\beta_0^j \gets \mathbf{0}_d$. {\hfill\textcolor{blue}{\%Initialization\%}}
        \FOR{$k=1, \cdots, K$}
            \STATE \quad Sample a subset of index $I_{k} \subseteq \{1,\cdots, n_{\text{train}}\}$ with batch size $|I_k| = m$.
            \STATE \quad 
                $ \beta^j_{k} \gets \beta^j_{k-1} + \frac{\eta}{m}\sum_{i \in I_{k}} (\bY^j_{i} - \bx_{i}^{\top} \beta^j_{k-1} ) \bx_i . $
            \STATE \quad $\hat{\beta}^{j}_{k} \gets {\beta^j_{k}}/{\|\beta^j_{k}\|_{2}}$  \hfill\textcolor{blue}{\%Normalization\%}
        \ENDFOR
    \ENDFOR
    \medskip
    \STATE {\textcolor{blue}{\%The Projection Matrix based on the Approximated Gradient Flow\%}}
    \STATE The projection matrix: $\hat{\boldsymbol{\beta}}_{k} = [\hat{\beta}^{1}_{k}, \hat{\beta}^{2}_{k}, \cdots, \hat{\beta}^{d'}_{k}] \in \mathbb{R}^{d \times d'}$ for $k=1, \cdots, K$. \\
    (Where each $\hat{\boldsymbol{\beta}}_{k}$ corresponds to some $\hat{\boldsymbol{\beta}}(\lambda)$.)
    \medskip
    \STATE {\textcolor{blue}{\%The Dimension-reduced Data Matrix Flow\%}}
    \STATE The dimension-reduced data matrix flow: 
    \STATE \quad $\tilde{\bX}_{\text{train}}(k) = \bX_{\text{train}} \hat{\boldsymbol{\beta}}_{k}$, for $k=1, \cdots, K$.
    \STATE \quad $\tilde{\bX}_{\text{val}}(k) = \bX_{\text{val}} \hat{\boldsymbol{\beta}}_{k}$, for $k=1, \cdots, K$.
    \medskip
    \STATE {\textcolor{blue}{\%Model selection based on the \texttt{AgFlow} matrix\%}}
    \FOR{$k=1, \cdots, K$ }
        \STATE Fit the target learner $h(k) = \mathtt{model}\left( \bX_{\text{train}} \hat{\boldsymbol{\beta}}_{k} \right)$ using the dimension-reduced data.
        \STATE Get the evaluation function $ \mathtt{Eval}_{k} = \mathtt{evaluator} \left(\bX_{\text{val}} \hat{\boldsymbol{\beta}}_{k}, h(k) \right)$.
    \ENDFOR
    \medskip
    \RETURN{ 
    $\hat{\boldsymbol{\beta}}^{*} = \mathop{\arg\max}_{\hat{\boldsymbol{\beta}}_{k}} \mathtt{evaluator} \left(\bX_{\text{val}} \hat{\boldsymbol{\beta}}_{k}, h(k) \right)$, for $k=1, \cdots, K$, as well as the optimal $k^{*}$ which corresponds to the optimal $\lambda^{*}$ under $\lambda\propto\frac{1}{k \sqrt{\eta}}$.}
\end{algorithmic}
\caption{\texttt{AgFlow} Algorithm for Model Selection.}
\label{alg:algorithm1}
\end{algorithm}

\subsection{Model Selection for $\ell_2$-penalized PCA over Approximated Gradient Flow}
In this section, we present the design of \texttt{AgFlow} algorithm (\textbf{Algorithm~\ref{alg:algorithm1}}) for obtaining the whole path of the loadings corresponding to each principal component for $\ell_2$-penalized PCA.  
Consider the $j^{th}$ principal component. Let $\bY^j$ be the $j^{th}$ principal subspace, which can be approximated by the Quasi-Principal Subspace Estimation Algorithm (\texttt{QuasiPS}) through the call of $\mathtt{QuasiPS(\bX,j)}$ (\textbf{Algorithm~\ref{alg:algorithm2}}).
The path of $\ell_2$-penalized PCA should be the solution path of Ridge regression in Eq.~(\ref{eq:pca-ridge}) with varying $\lambda$ from $0\to\infty$. 

With the implicit regularization of Stochastic Gradient Descent (SGD) for Ordinary Least Squares (OLS)~\citep{ali2020implicit}, the solution path is equivalent to the optimization path of the following OLS estimator using SGD with zero initialization, such that
\begin{equation}\label{eq:pca-ols} 
    \mathop{\min}_{\beta \in \mathbb{R}^{d}}\
    \frac{1}{2n} \| \bY^j - \bX \beta \|_2^2\ . 
\end{equation}
More specifically, with a constant step size $\eta > 0$, an initialization $\beta^j_0 = \mathbf{0}_d$, and mini-batch size $m$, every SGD iteration updates the estimation as follows, 
\begin{align} \label{sgdpca-sol}
    \beta^{j}_{k} &= \beta^{j}_{k-1} + \frac{\eta}{m} \cdot
    \sum_{i \in I_{k}} (\bY_i^j - \bX_i^{\top}\beta^{j}_{k-1} )  \mathbf{x}_i \ ,
\end{align}
for $k=1,2,\cdots, K$, and thus the solutions on the (stochastic) gradient flow path for the $j^{th}$ principal component can be obtained.
According to~\citep{ali2020implicit}, the relationship of the explicit regularization $\lambda$ and the implicit regularization effects introduced by SGD is $\lambda\propto\frac{1}{k \sqrt{\eta}}$. Thus, with the total number of iteration steps $K$ large enough, the proposed algorithm could compete the path of penalized PCA for a full range of $\lambda$, but with a much lower computation cost.

Since in the problem of model selection of $\ell_2$-penalized PCA based on Ridge estimator, we need to select the optimal $\hat\beta(\lambda^{*})$ corresponding to the the optimal $\lambda^{*}$. Here we deal with the same model selection problem but with an alternative algorithm which uses the \texttt{AgFlow} algorithm instead of using matrix inverse in Ridge estimator. Therefore, we need to select the optimal $\hat\beta^{*}$ corresponding to the the optimal $k^{*}$ with $k^{*} \propto \frac{1}{\lambda^{*}\sqrt{\eta}}$. 
To obtain the optimal $\lambda^{*}$, $k$-fold cross-validation is usually applied on a searching grid of $\lambda$-s in the model selection. As an analog of obtaining the optimal $k^{*}$, the proposed \texttt{AgFlow} algorithm is firstly used to get the iterated projection vector $\hat\beta_{k}$ of the given training data, which corresponds to some $\hat{\beta}(\lambda^{*})$ with $k \propto \frac{1}{\lambda \sqrt{\eta}}$, then to select the optimal $\beta^{*}$ based on the performance on the validation data.



Finally, \textbf{Algorithm.~\ref{alg:algorithm1}} outputs the best projection matrix $\hat{\boldsymbol{\beta}}_{k^*} \in \mathbb{R}^{d \times d'}$, which maximizes the evaluator $\mathtt{Eval}_{k} = \mathtt{evaluator} \left(\bX_{\text{val}} \hat{\boldsymbol{\beta}}_{k}, h(k) \right)$, for $k=1, \cdots, K$. 
Where the index $k \propto \frac{1}{\lambda \sqrt{\eta}}$, and each column of the projection matrix $\hat{\boldsymbol{\beta}}_{k}$ is a normalized projection vector $\|\hat{\beta}^{j}_{k}\|_2=1$. Note that, as discussed in the preliminaries in Section~\ref{sec:prelim}, when the sample covariance matrix $1/n\bX^\top\bX$ is non-singular (when $n\gg d$), there is no need to place any penalty here, 
i.e., $\lambda \to 0$, and $k \to \infty $, as the normalization would remove the effect of the $\ell_2$-regularization~\citep{zou2006sparse} considering Karush–Kuhn–Tucker conditions. However, when $d\gg n$, the sample covariance matrix $1/n\bX^\top\bX$ becomes singular, and the Ridge-liked estimator starts to shrink the covariance matrix as in Eq.~(\ref{eq:ridge-closedform}), i.e., $\lambda \neq 0$, and $k$ is some finite integer but not $\infty$, making the sample covariance matrix invertible and the results penalized in a covariance-regularization fashion~\citep{witten2009covariance}. Even though the normalization would rescale the vectors to a $\ell_2$-ball, the regularization effect still remains.

\subsection{Near-Optimal Initialization for Quasi-Principal Subspace}

The goal of the \texttt{QuasiPS} algorithm is to approximate the principal subspace of PCA with given data matrix with extremely low cost, and \texttt{AgFlow} would fine-tune the rough  quasi-principal projection estimation (i.e. the loadings) and obtain the complete path of the $\ell_2$-penalized PCA accordingly. 
While there are various low-complexity algorithms in this area
, such as~\citep{hardt2014noisy,shamir2015stochastic,de2015global,balsubramani2013fast}, 
%
we derive the Quasi-Principal Subspace (\texttt{QuasiPS}) estimator (in \textbf{Algorithm}~\ref{alg:algorithm2}) using the stochastic algorithms proposed in~\citep{shamir2015stochastic}. More specifically, \textbf{Algorithm}~\ref{alg:algorithm2} first pursues a rough estimation of the $j^{th}$ principal component projection (denoted as $\tilde{w}_L$ after $L$ iterations) using the stochastic approximation~\citep{shamir2015stochastic}, then 
obtains the quasi-principal subspace $\bY^j$ through projection $\bY^j=\bX \tilde{w}_L$.

Note that $\tilde{w}_L$ is not a precise estimation of the loadings corresponding to its principal component (compared to our algorithm and~\citep{oja1985stochastic} etc.), however it can provide a close solution in an extremely low cost. In this way, we consider $\bY^j=\bX \tilde{w}_L$ as a reasonable estimate of the principal subspace. 
%
%
With a random unit initialization $\tilde{w}_{0}$, $\tilde{w}_{L}$ converges to the true principal projection $\mathbf{v}^*$ in a fast rate under mild conditions, even when $\tilde{w}_{0}^\top\mathbf{v}^* \geq \frac{1}{\sqrt{2}}$~\citep{shamir2015stochastic}. Thus, our setting should be non-trivial.


\begin{algorithm}[H]
\algsetup{linenosize=\normalsize}
\begin{algorithmic}
    \STATE \textbf{Input:} 
        \STATE \quad Data matrix $\bX = \{\mathbf{x}_1, \cdots ,\mathbf{x}_n\}$, index $j$.
    \STATE \textbf{Parameters:} 
        \STATE \quad Step size $\eta_{2}$, epoch length $M$, iterations $L$.
    \STATE \textbf{Output: } The $j^{th}$ Quasi-PS, $\bY^j$.
    \medskip
    \STATE \textbf{Initialization: } $\check{w}_{0}\sim \mathcal{N}(0,\mathbf{I}_d)$ \text{ and } $\tilde{w}_{0}\gets \frac{\check{w}_{0}}{\|\check{w}_{0}\|_2}$.
    \hfill \textcolor{blue}{\%Random Unit Initialization.\%}
    \STATE \textbf{for} $l=1,\cdots, L$ \textbf{do}
        \STATE \quad Let $ \tilde{u} = \frac{1}{n}\sum_{i=1}^{n} \mathbf{x}_i (\mathbf{x}_i^{\top} \tilde{w}_{l-1})$.
        \STATE \quad $w_{0} = \tilde{w}_{l-1} $.
        \STATE \quad \textbf{for} $s=1,2,\cdots, M$ \textbf{do}
            \STATE \quad \quad Pick $i_s \in \{1,\cdots, n\}$ uniformly at random.
            \STATE \quad \quad 
                $w_s^{'} = w_{s-1} + \eta_{2} [\mathbf{x}_{i_s} (\mathbf{x}_{i_s}^{\top} w_{s-1} - \mathbf{x}_{i_s}^{\top} \tilde{w}_{l-1}) + \tilde{u} ] . $
            \STATE \quad \quad
             $w_{s} = \frac{w_s^{'}}{\|w_s^{'}\|_2}$
        \STATE \quad \textbf{end for}
        \STATE \quad $\tilde{w}_{l} = w_{M}$.
    \STATE{\textbf{end for}}
    \medskip
    \STATE  $\bY^j \gets \bX \tilde{w}_{L}$ \hfill\textcolor{blue}{\%Quasi Principal Subspace\%}
    \RETURN{$\bY^j$.}
 \end{algorithmic}
\caption{$\mathtt{QuasiPS(\bX,j)}$ -- Quasi-Principal Subspace Approximation. 
}
\label{alg:algorithm2}
\end{algorithm}




\subsection{Algorithm Analysis }
In this section, we analyze the proposed algorithm from perspectives of statistical performance and its computational complexity.

\subsubsection{Statistical Performance}

\texttt{AgFlow} algorithm consists of two steps: Quasi-PS initialization and solution path retrieval. As the goal of our research is fast model selection on the complete solution path for $\ell_2$-penalized PCA over varying penalties, the performance analysis of the proposed algorithm can be decomposed into two parts.

\noindent{\emph{\textbf{Approximation of Quasi-PS to the true principal subspace.}}} In \texttt{Algorithm.~\ref{alg:algorithm2}}, the $\mathtt{QuasiPS}$ algorithm first obtains a Quasi-PS projection $\tilde{w}_L$ using $L$ epochs of low-complexity stochastic approximation, then it projects the sample $\bx$ to get the $j^{th}$ principal subspace  $\bY^{j}$ via $\bx \tilde{w}_L$. 

\begin{lemma}
Under some mild conditions as in~\cite{shamir2015stochastic} and given the true principal projection $w^*$, with probability at least $1-\mathrm{log}_{2}(1/\varepsilon)\delta$, the the distance between $w^*$ and $\tilde{w}_L$ holds that
\begin{equation}\label{eq:init-error}
    \| \tilde{w}_L-w^*\|_2^2\leq 2-2\sqrt{1-\varepsilon} ,
\end{equation}
provided that $L=\mathrm{log}(1/\varepsilon)/\mathrm{log}(2/\delta)$.
\end{lemma}
It can be easily derived from (Theorem~1. in~\cite{shamir2015stochastic}). When $\varepsilon\to 0$, $\|\tilde{w}_L-w^*\|_2^2\to 0$ and the error bound becomes tight. Suppose samples in $\bX$ are i.i.d. realizations from the random variable $X$ and $\mathbb{E}XX^\top=\Sigma^*$ denote the true covariance.

\begin{theorem}
Under some mild conditions as in~\cite{shamir2015stochastic}, the distance between the Quasi-PS and the true principal subspace holds that
  \begin{equation}
  \begin{aligned}
    \underset{\bx\sim{X}}{\mathbb{E}}\|\mathbf{x}\tilde{w}_L-\mathbf{x}w^*\|_2^2&= (\tilde{w}_L-w^*)^\top \Sigma^* (\tilde{w}_L-w^*)\\
    &\leq \lambda_\mathrm{max}(\Sigma^*)\|\tilde{w}_L-w^*\|_2^2\\
    &= (2-2\sqrt{1-\varepsilon})\cdot\lambda_\mathrm{max}(\Sigma^*)\ ,
    \end{aligned}
\end{equation}
where $\lambda_\mathrm{max}(\cdot)$ refers to the largest eigenvalue of a matrix. 
\end{theorem}
When considering the largest eigenvalue $\lambda_\mathrm{max}(\Sigma^*)$ as a constant, Quasi-PS is believed to achieve exponential coverage rate for principal subspace approximation for every sample. Thus, the statistical performance of Quasi-PS can be guaranteed.

\noindent{\emph{\textbf{Approximation of Approximated Stochastic Gradient Flow to the Solution Path of Ridge.}}} In~\citep{ali2019continuous,ali2020implicit}, the authors have demonstrated that when the learning rate $\eta\to 0$, the discrete-time SGD and GD algorithms would diffuse to two continuous-time dynamics over (stochastic) gradient flows, i.e., $\hat{\beta}_\mathrm{sgf}(t)$ and $\hat{\beta}_\mathrm{gf}(t)$ over continuous time $t>0$. According to Theorem 1 in~\cite{ali2019continuous}, the statistical risk between Ridge and continuous-time gradient flow is bounded by 
\begin{equation}
    \mathrm{Risk}(\hat{\beta}_\mathrm{gf}(t),\beta^*)\leq 1.6862 \cdot \mathrm{Risk}(\widehat\beta_{\text{ridge}}(1/t), \beta^*)
\end{equation}
where $\mathrm{Risk}(\beta_1,\beta_2)=\mathbb{E}\|\beta_1-\beta_2\|^2_2$, $\beta^*$ refers to the true estimator, and $\lambda=1/t$ for Ridge. While the stochastic gradient flow enjoys a faster convergence but with slightly larger statistical risk, such that 
\begin{equation}
    \begin{aligned}
    \mathrm{Risk}(\hat{\beta}_\mathrm{sgf}(t),\beta^*)\leq \mathrm{Risk}(\hat{\beta}_\mathrm{gf}(t),\beta^*)+{o}\left(\frac{n}{m}\right)\ ,
    \end{aligned}
\end{equation}
where $m$ refers to the batch size and ${o}\left(\frac{n}{m}\right)$ is an error term caused by the stochastic gradient noises. 
Under mild conditions, with discretization ($\mathbf{d} t=\sqrt{\eta}$), we consider the $k^{th}$ iteration of SGD for Ordinary Least Squares, denoted as $\beta_k$, which tightly approximates to $\hat{\beta}_\mathrm{sgf}(t)$ in a ${o}(\sqrt{\eta})$-approximation with $t=k\sqrt{\eta}$. 
In this way, given the learning rate $\eta$ and the total number of iterations $K$, the implicit Ridge-like \texttt{AgFlow} screens the $\ell_2$-penalized PCA with varying $\lambda$ in the range of
\begin{equation}
    \frac{1}{K\sqrt{\eta}}\leq \lambda\leq \frac{1}{\sqrt{\eta}}\ ,
\end{equation}
with bounded error in both statistics and approximation.

In this way, we could conclude that under mild conditions, \texttt{QuasiPS} can well approximate the true principal subspace ($d'\ll n$) while \texttt{AgFlow} retrieves
a tight approximation of the Ridge solution path..

\subsubsection{Computational Complexity} The proposed algorithm consists of two steps: the initialization of the quasi-principal subspace and the path retrieval. To obtain a fine estimate of Quasi-PS and hit the error in Eq.~(\ref{eq:init-error}), one should run Shamir's algorithm~\citep{shamir2015stochastic} with $$O\left((\mathrm{rank}(\Sigma^*)/\mathrm{eigengap}(\Sigma^*))^2\mathrm{log}(1/\varepsilon)\right)$$ iterations, where $\mathrm{rank}(\cdot)$ refers to the matrix rank, $\mathrm{eigengap}(\cdot)$ refers to the gap between the first and second eigenvalues, and $\varepsilon$ has been defined in \textbf{Lemma 2} referring to the error of principal subspace estimation.

Furthermore, to get the loadings corresponding to the $j^{th}$ principal subspace, \texttt{AgFlow} uses $K$ iterations for OLS to obtain the estimate of $K$ models for $\ell_2$-penalized PCA, where each iteration only consumes $O(m\cdot d^2)$ complexity with batch size $m$, which gets total $O(K\cdot m \cdot d^2)$ for $K$ models, and total $O(d' \cdot K\cdot m \cdot d^2)$ with the reduced-dimension $d'$. 
Moreover, we also propose to run $\texttt{AgFlow}$ with full-batch size $m=n$ using gradient descent per iteration, which only consumes $O(d^2)$ per iteration with lazy evaluation of $\bX^\top\bX$ and $\bX^\top\bY$, with total $O(K\cdot d^2)$ for $K$ models, which gets $O(d' \cdot K \cdot \cdot d^2)$ with the reduced-dimension $d'$. 

To further improve \texttt{AgFlow} without incorporating higher-order complexity, we carry out the experiments by running a mini-batch \texttt{AgFlow}, and a full-batch \texttt{AgFlow} (i.e., $m=n$) with lazy evaluation of $\bX^\top\bX$ and $\bX^\top\bY$ in parallel for model selection.



\section{Experiments}\label{sec:experiments}

In this section, we show some experiments on real-world datasets with a significantly large number of features; that fits well in the natural High Dimension Low Sample Size (HDLSS) settings.
Since cancer classification has remained a great challenge to researchers in microarray technology, we try to adopt our new algorithm on these gene expression datasets.
In particular, except for three publicly available gene expression datasets~\citep{zhu2007markov}, the well-known FACES dataset~\citep{huang2008labeled} in machine learning is also considered in our study. A brief overview of these four datasets is summarized in Table~\ref{table:dataset}.


\begin{table}[htbp]\centering
\caption{Description of the FACES dataset and Three Microarray Datasets (i.e. gene expression dataset of Colon Tumor, ALL-AML Leukemia, Central Nervous System).}
\label{table:dataset}
\begin{tabular}{lccc}\toprule
Dataset & $\#$Total Features ($d$) & $\#$Samples ($n$) & $\#$Classes \\
\midrule
FACES & 4096 & 400 & 10 \\
Colon Tumor & 2000 & 62 & 2 \\
ALL-AML & 7129 & 72 & 2 \\
Central Nervous System & 7129 & 60 & 2 \\
\bottomrule
\end{tabular}
\end{table}

\begin{figure*}[htbp]
\centering
\includegraphics[width=0.9\textwidth]{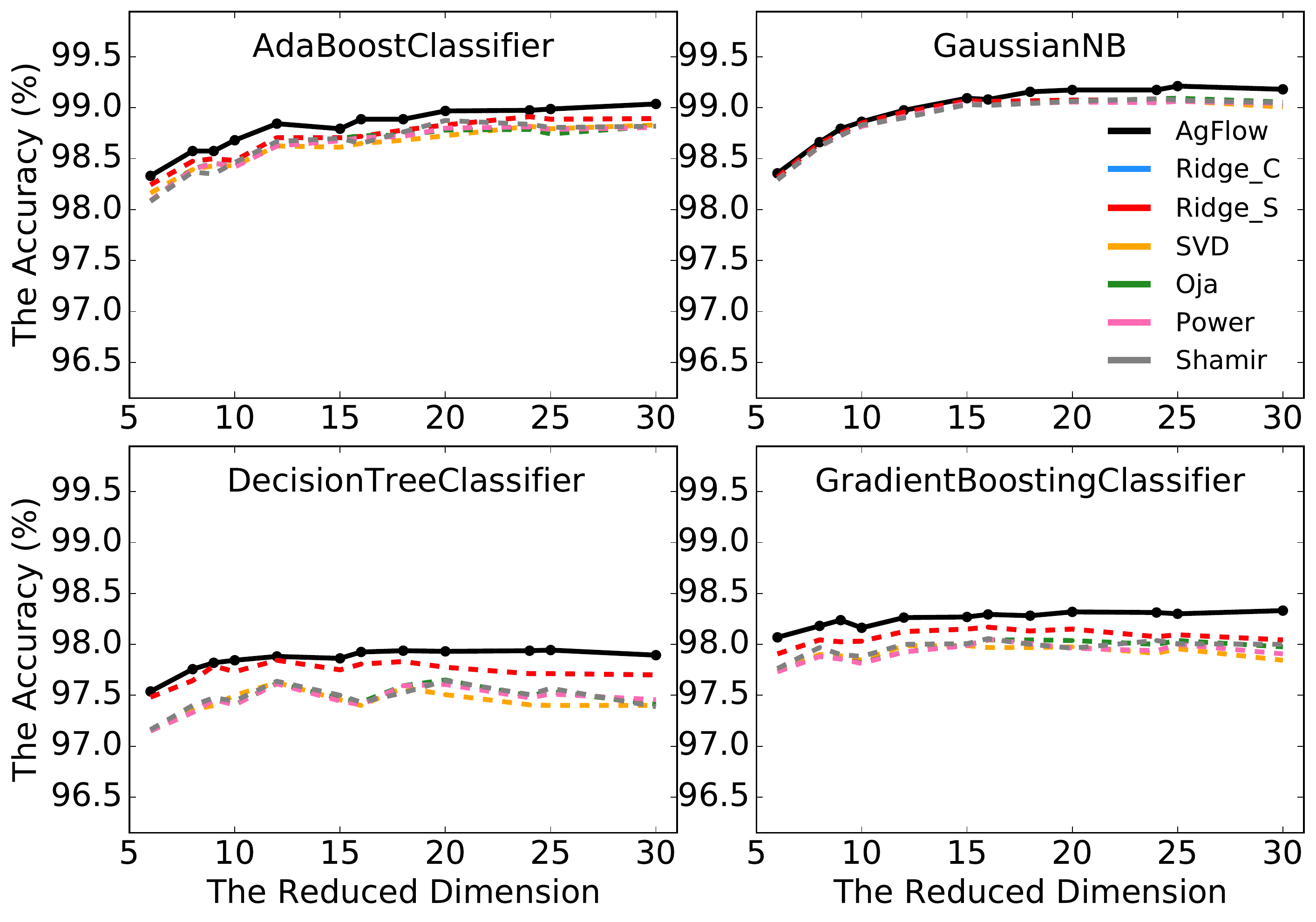}
\caption{Performance Comparisons on Dimension Reduction between \texttt{AgFlow} and Other Baseline Algorithms based on the Validation Accuracy of Adaptive Boosting Classifier, Gaussian Naive Bayes, Decision Tree Classifier, Gradient Boosting Classifier on FACES Dataset, respectively. SVD: Vanilla PCA based on SVD~\citep{jolliffe1986principal}, Oja: Oja's stochastic PCA method~\cite{oja1985stochastic}, Power: Power Iteration method~\citep{gene1996matrix}, Shamir: Shamir's Variance Reduction method~\citep{shamir2015stochastic} and Ridge: Ridge-based PCA~\citep{zou2006sparse}.
(\emph{Ridge\_C stands for the ridge estimator based on the closed-form as in Eq.~(\ref{eq:ridge-closedform}), Ridge\_S stands for the ridge estimator based on scikit-learn solvers.})
}
\label{fig:overall-one}
\end{figure*}

\begin{figure*}[!htb]
\centering
\subfloat[Validation and Test on FACES ]{\includegraphics[width=0.48\textwidth]{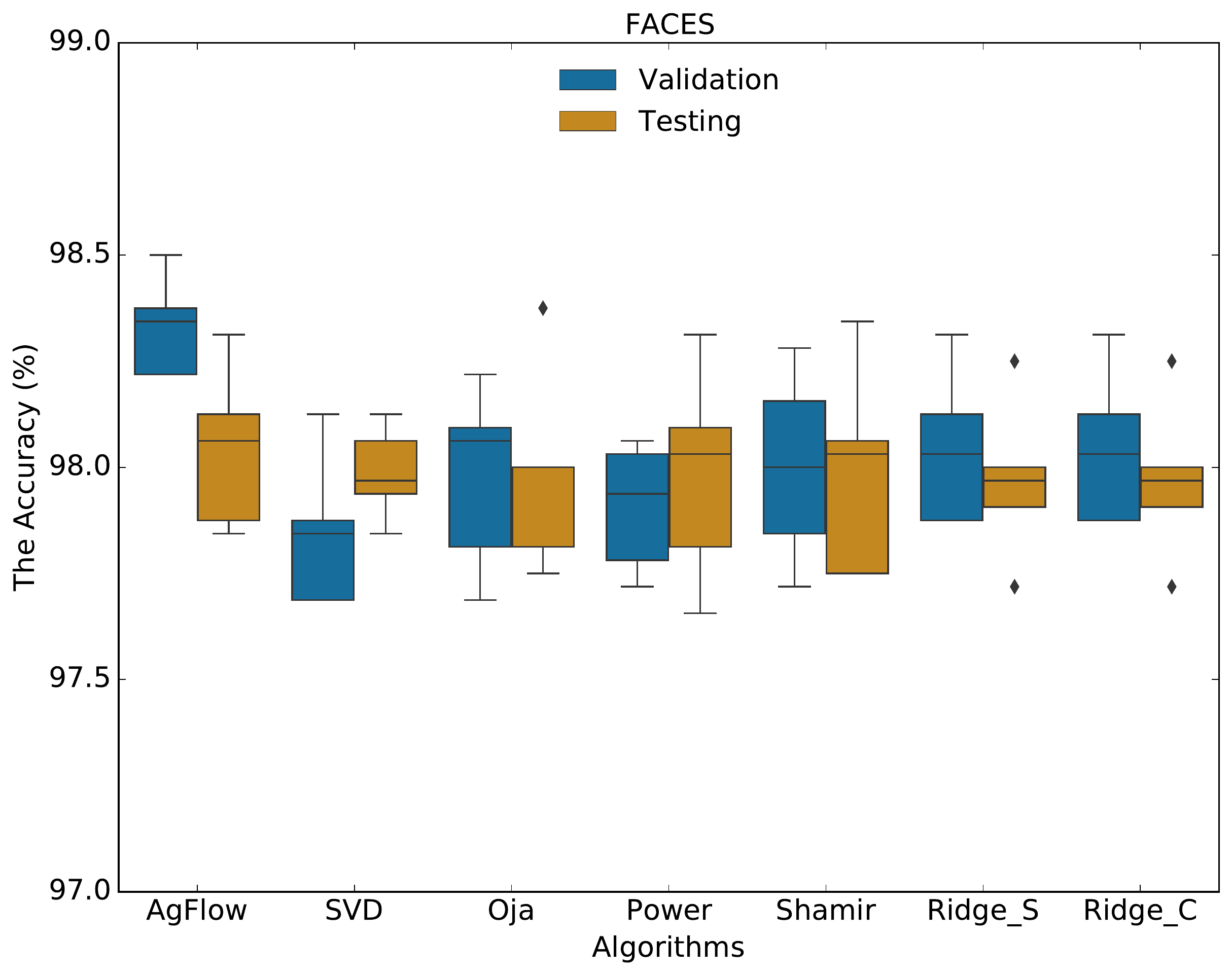}}\
\subfloat[Validation and Test on Colon Tumor]{\includegraphics[width=0.48\textwidth]{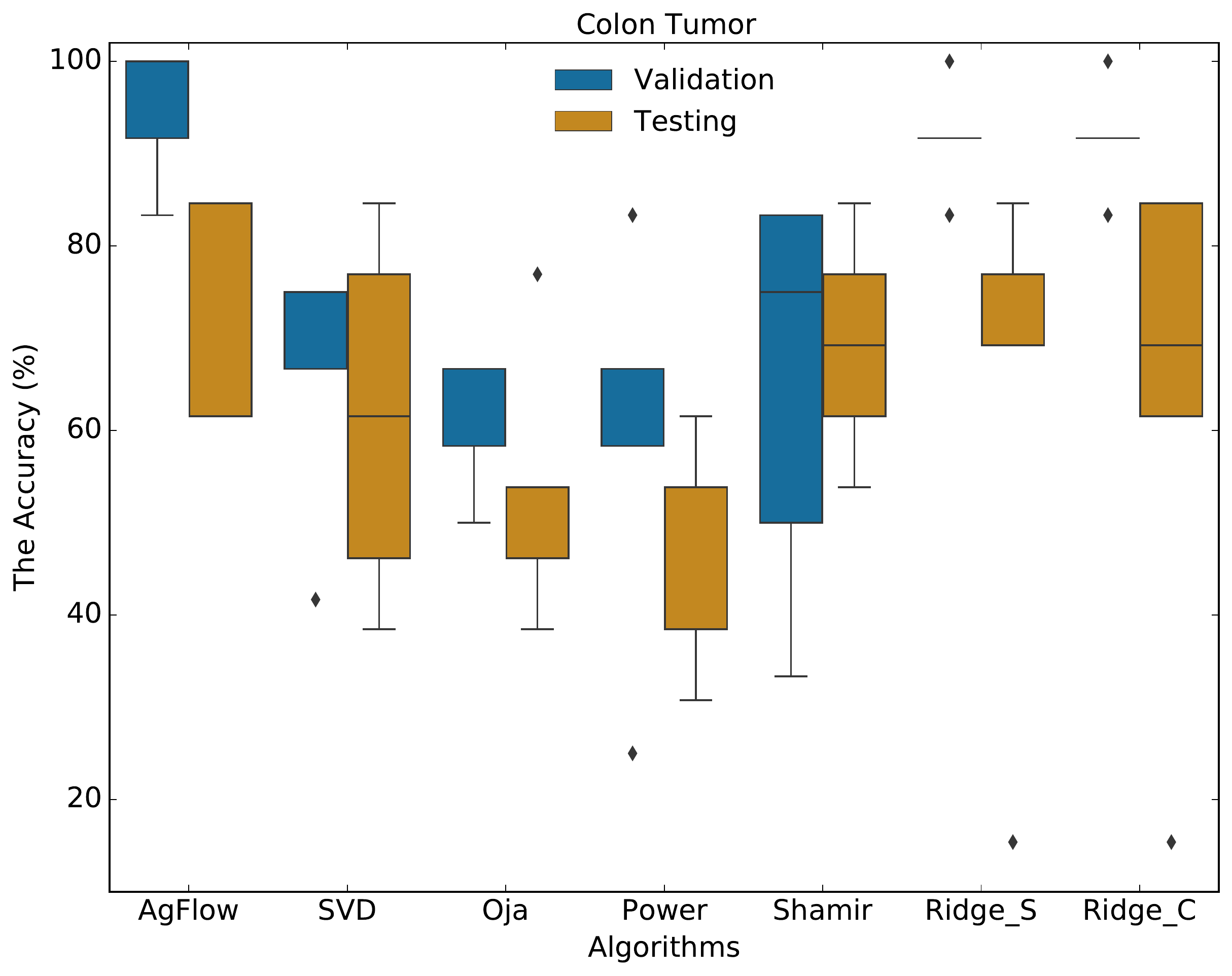}}\\
\subfloat[Validation and Test on ALL-AML ]{\includegraphics[width=0.48\textwidth]{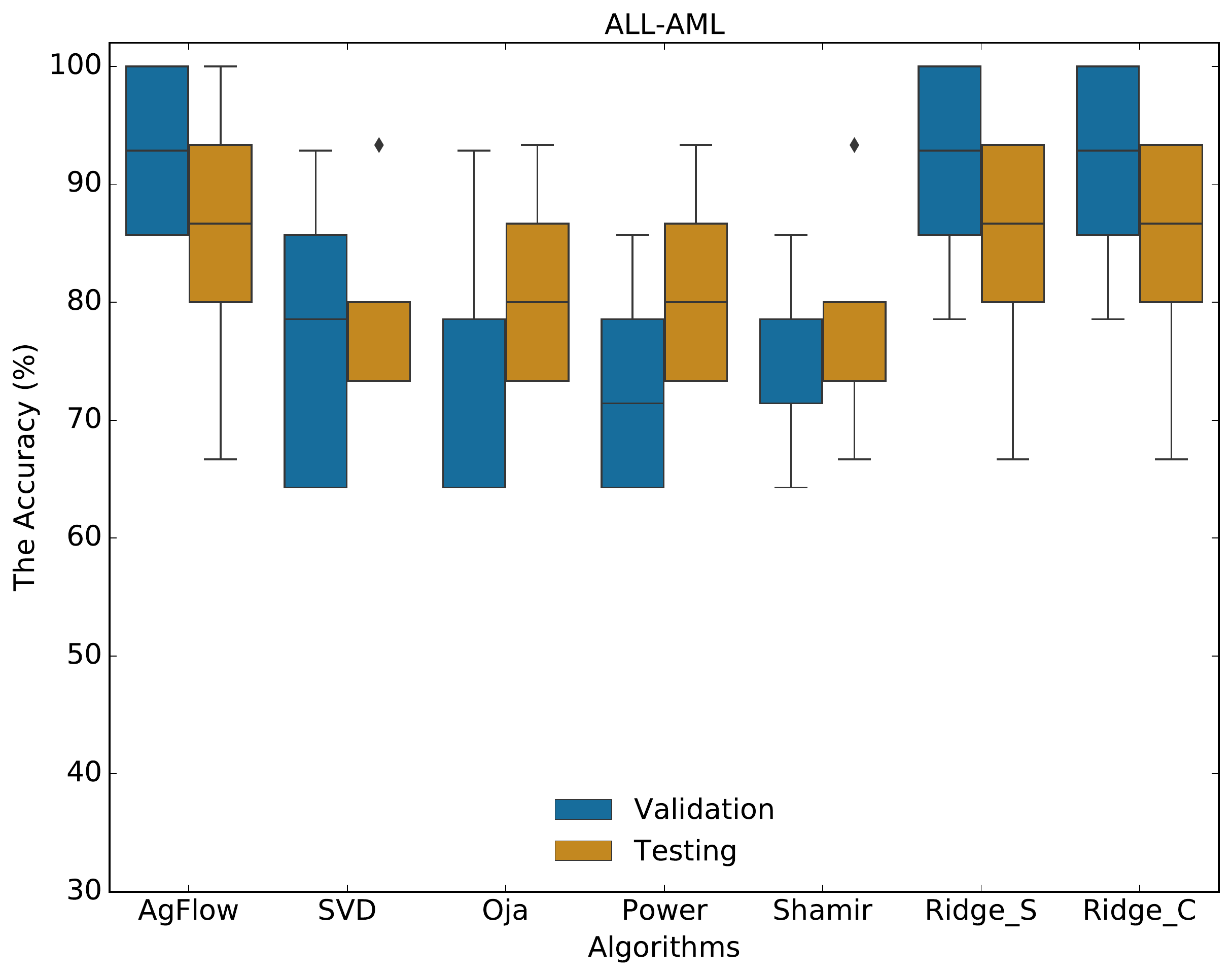}}\
\subfloat[Validation and Test on Central Nervous System ]{\includegraphics[width=0.48\textwidth]{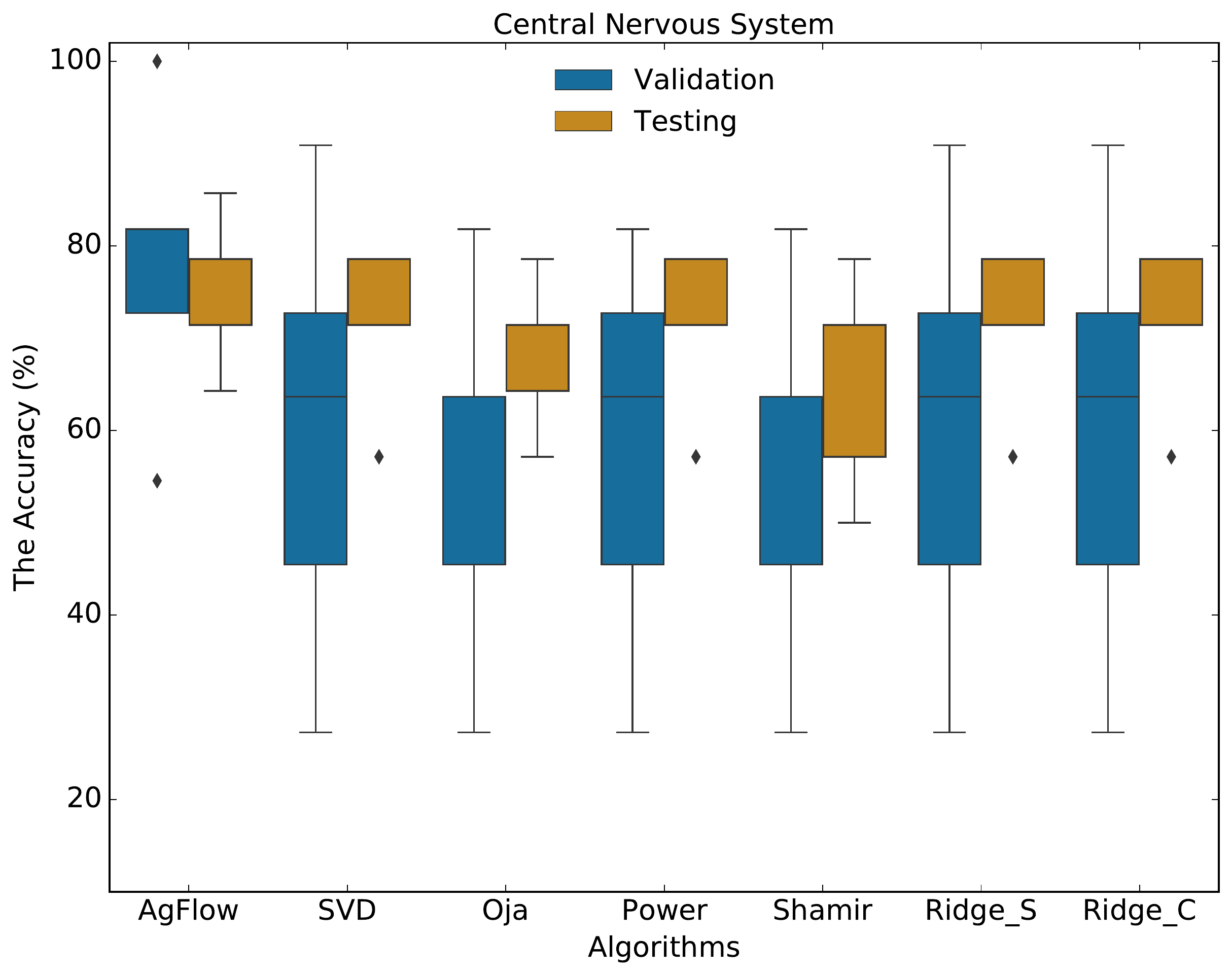}}\\
\caption{Performance Comparisons of Validation and Testing Accuracy of Different Dimension Reduction Methods, \texttt{AgFlow} and Other Baseline Algorithms, on FACES (with $d'=30$ and Gradient Boosting Classifier), Colon Tumor (with $d'=30$ and Quadratic Discriminant Analysis), ALL-AML (with $d'=30$ and Random Forest Classifier), Central Nervous System Dataset (with $d'=24$ and Gradient Boosting Classifier). 
SVD: vanilla PCA based on SVD~\citep{jolliffe1986principal},
Oja: Oja's Stochastic PCA method~\cite{oja1985stochastic}, Power: Power Iteration method~\citep{gene1996matrix}, Shamir: Shamir's Variance Reduction method~\citep{shamir2015stochastic} and Ridge: Ridge-based PCA~\citep{zou2006sparse}.
(\emph{Ridge\_C stands for the ridge estimator based on the closed-form as in Eq.~(\ref{eq:ridge-closedform}), Ridge\_S stands for the ridge estimator based on scikit-learn solvers.})
}
\label{fig:4data}
\end{figure*}

\subsection{Experiment Setups}

\paragraph{\textbf{Evaluation procedure of the \texttt{AgFlow} algorithm.}} 
There are two regimes to demonstrate the performance of the proposed model selection method; the first is to evaluate the accuracy of the \texttt{AgFlow} algorithm based on $k$-fold cross-validation, which we call it evaluation-based model selection; the second is to do prediction based on the given training-validation-testing set which consists of three steps, i.e., model selection, model evaluation and prediction, which we call it prediction-based model selection. Usually, in the real-world applications, the prediction-based model selection is used, where the testing set is unseen in advance.
The proceeding step would be to split the raw data into training-validation set for further cross-validation in evaluation-based model selection and training-validation-testing set for prediction-based model selection. There are two main steps, first is to get the the projection matrix of the the training data using the \texttt{AgFlow} algorithm; second is to apply the projection matrix to the validation/testing set.

Here we take the prediction-based model selection as an example.
To do model selection using the \texttt{AgFlow} algorithm, \emph{firstly} we need to get the projection matrix flow of the given training set by running the \texttt{AgFlow} algorithm, e.g. $\hat{\boldsymbol{\beta}}_{k} \in \mathbb{R}^{d \times d'}$ for $k=1,\cdots, K$. Then the dimension-reduced training-validation-testing data matrix flow can be obtained by matrix multiplication, e.g. $\tilde{\bX}_{\text{train}}(k) = \bX_{\text{train}} \hat{\boldsymbol{\beta}}_{k}$,
where $\hat{\boldsymbol{\beta}}_{k} = [\hat{\beta}^{1}_{k}, \hat{\beta}^{2}_{k}, \cdots, \hat{\beta}^{d'}_{k}] \in \mathbb{R}^{d \times d'}$. Each column $\hat{\beta}^{j}_{k}$ is the $j^{th}$ projection vector, i.e., the $j^{th}$ loadings corresponding to the $j^{th}$ principal component, which approximates the $\ell_2$-penalized PCA with the tuning parameter $\lambda$, under the calibration $\lambda \propto 1/(k \sqrt{\eta})$.
\emph{Then} the dimension-reduced training data matrix flow is fed into the target learner $h(k) = \mathtt{model}(\bX_{\text{train}} \hat{\boldsymbol{\beta}}_{k})$ for performance tuning which outputs models $h(k)$ for $k=1, \cdots, K$. \emph{Lastly}, the dimension-reduced validation data matrix flow is used to choose the optimal model with best performance according to the evaluator $\mathtt{evaluator}( \bX_{\text{val}} \hat{\boldsymbol{\beta}}_{k}, h(k))$ for $k=1,\cdots, K$, which gives the optimal $\hat{\boldsymbol{\beta}}^{*}=\mathop{\arg\max}_{\hat{\boldsymbol{\beta}}_{k}}\mathtt{evaluator}( \bX_{\text{val}} \hat{\boldsymbol{\beta}}_{k}, h(k))$ and the optimal $k^{*}$. 
Note that each data flow matrix possesses some implicit regularization introduced by the \texttt{AgFlow} algorithm, which corresponds to an explicit penalty in Ridge. Under the calibration $\lambda \propto 1/(k \sqrt{\eta})$, we have $\hat{\boldsymbol{\beta}}_{k}\approx  \hat{\boldsymbol{\beta}}(\lambda)$, $\hat{\boldsymbol{\beta}}^{*}\approx  \hat{\boldsymbol{\beta}}(\lambda^{*})$, with $\lambda^{*} \propto 1/(k^{*} \sqrt{\eta})$, thus we can do model selection using results based on \texttt{AgFlow} algorithm.

\paragraph{\textbf{Settings of the \texttt{AgFlow} algorithm. }}
\begin{itemize}
    \item \emph{Construction of training-validation-testing set. }
    For the above four datasets, we randomly split the raw data samples into training-validation-testing set with a fixed split ratio of $60\%-20\%-20\%$ within each class.
    Then the sample size for the training-validation-testing set is 
    $(240, 80, 80)$, $(37, 12, 13)$, $(43, 14, 15)$, $(35, 11, 14)$
    for FACES, Colon Tumor, ALL-AML Leukemia, Central Nervous System data, respectively. 
    Thus dimension/sample size ratio $d/n$ of the training set is $17.1$, $54.1$, $165.8$, $203.7$ accordingly. 
    \item \emph{Settings of default parameters.}
    For the default parameters in \texttt{AgFlow}, the number of iterations is set  $K=5000$, the step size $\eta = 0.5\times 10^{-4}$, the batch size $\min(100, n/2)$, and the reduced dimension $d'=30$.
    For the values of explicit regularization of $\lambda$ in \texttt{Ridge}, the $\ell_2$-penalized PCA, we take $100$ values in the log-scale ranging from $10^{-4}$ to $10^{4}$ as the searching grid.
    For the default parameters in $\texttt{QuasiPS}(\bX, j)$, we take the same default values as those specified in the original paper~\cite{shamir2015stochastic}, where the step size $\eta_2 = \frac{1}{\bar{r}{n}}$, and $\bar{r} = \frac{1}{n}\sum_{i=1}^{n} \|\bx_i\|_2^2$, the epoch length $M=n$, and the number of iterations $L=100$.
\end{itemize}



\paragraph{\textbf{Baseline PCA Algorithms. }} To demonstrate the performance of the \texttt{AgFlow} algorithm, we compare the results with some other comparable methods, such as Oja's method~\cite{oja1985stochastic}, Power iteration~\citep{gene1996matrix}, Shamir's Variance Reduction method~\citep{shamir2015stochastic}, vanilla PCA~\citep{jolliffe1986principal}, and Ridge-based PCA~\citep{zou2006sparse} (two variants: the closed-form ridge estimator in Eq.~(\ref{eq:ridge-closedform}), Ridge\_C, and that based on scikit-learn solvers, Ridge\_S).

\subsection{Overall Comparisons of Model Selection}
In this section, we evaluate the performance of the proposed \texttt{AgFlow} algorithm and compare it with other baseline algorithms (especially in the performance comparisons with Ridge-based estimator) using FACES data, and three gene expression data of Colon Tumor, ALL-AML Leukemia, and Central Nervous System, respectively. In all these experiments, the training datasets have a limited number of samples and a significantly large number of features in the dimension reduction problem. For example, $d/n$ ranges from $10$ to $120$, which is significantly larger than one in the four datasets. The common learning problem becomes ill-posed and models are all over-fit to the small training datasets. Model selection with the validation set becomes a crucial issue to improve the performance.

Fig.~\ref{fig:overall-one} presents the overall performance comparisons on the dimension reduction problem between \texttt{AgFlow} and other baseline algorithms using FACES dataset, where the classification accuracy with dimension-reduced data is used as the metric. As only \texttt{AgFlow} and Ridge are capable of estimating penalized PCA models for model selection, in Fig.~\ref{fig:overall-one}, we select the best models of both \texttt{AgFlow} and Ridge in terms of validation accuracy. For a fair comparison, we compare \texttt{AgFlow} with Ridge for model selection in a similar range of penalties ($\lambda$) using a similar budget of computation time, while we make sure that the time spent by \texttt{AgFlow} algorithm is much shorter than Ridge (Please refer Table~\ref{table:timeused} for the time consumption comparisons between \texttt{AgFlow} and Ridge.).

Under such critical HDLSS settings, usually all algorithms work poorly while \texttt{AgFlow} outperforms all these algorithms in most cases. Furthermore, Shamir's~\citep{shamir2015stochastic} method, Oja's method~\citep{oja1985stochastic}, Power iteration method and the vanilla PCA based on SVD, all achieve the similar performance in these settings, it seems these algorithms beat the best performance achievable for the unbiased PCA estimator without any regularization under ill-posed and HDLSS settings. The comparison between \texttt{AgFlow} and unbiased PCA estimators demonstrates the performance improvement contributed by the implicit regularization effects~\citep{ali2020implicit} and the potentials of model selection with validation accuracy. Furthermore, the comparison between \texttt{AgFlow} and Ridge indicates that the implicit regularization effect of SGD provides the model estimator with higher stability than Ridge in estimating penalized PCA under HDLSS settings, as the matrix inverse used in Ridge is unstable when the model is ill-posed~\citep{haber2008numerical}. Furthermore, the continuous trace of SGD provides model selector with more flexibility than Ridge in screening massive models under varying penalties with fine-grained granularity. 

Fig.~\ref{fig:4data} gives the performance comparison of validation and testing accuracy of different dimension reduction methods on different datasets, including \texttt{AgFlow} and other baseline algorithms such as vanilla PCA based on SVD~\citep{jolliffe1986principal}, 
Oja's Stochastic PCA method~\cite{oja1985stochastic}, Power Iteration method~\citep{gene1996matrix}, Shamir's Variance Reduction method~\citep{shamir2015stochastic}, and Ridge: Ridge-based PCA~\citep{zou2006sparse}. 
Fig.~\ref{fig:4data} shows that for the gene expression dataset of the Colon Tumor and Central Nervous System, the \texttt{AgFlow} algorithm outperforms other baseline algorithms with an overwhelming improvement with respect to the validation accuracy as well as the testing accuracy.
For the FACES dataset, not much advantage of \texttt{AgFlow} is gained because all the algorithms achieve an accuracy above $90\%$, thus the improvement is less than $5\%$.
For the ALL-AML dataset, the performance of all the algorithms varies a lot, our \texttt{AgFlow} is still the best one with respect to the validation accuracy, however, it is not the case when applied to the testing accuracy. The reason may be that, with one shot of training-validation-testing splitting, there is some variability in the data splitting and as the sample size is not that large that makes this uncertainty worse, which also explains that the testing accuracy is somewhat larger than the validation accuracy for some algorithms.

In this way, based on the comparisons of different dimension reduction methods using the same data with a given classifier function as in Fig.~\ref{fig:overall-one} and the the comparisons using different datasets Fig.~\ref{fig:4data}, we can conclude that \texttt{AgFlow} is more effective than Ridge for estimating massive models and selecting the best models for penalized PCA, with the same or even stricter budget conditions. 
We also present the comparison results based on different datasets in Fig.~\ref{fig:4data} using various classifiers. Similar results are obtained: Ridge works well as more samples provided and \texttt{AgFlow} outperforms Ridge estimator in most cases.


\begin{table}[ht]
\centering
\scriptsize
\caption{Comparison of Time Consumption/(\#Models) in Seconds using FACES dataset, and gene expression data for Colon Tumor, ALL-AML Leukemia, Central Nervous System. 
(\emph{Ridge\_C stands for the ridge estimator based on the closed-form as in Eq.~(\ref{eq:ridge-closedform}), Ridge\_S stands for the ridge estimator based on scikit-learn solvers.})
}
\label{table:timeused}
\begin{tabular}{llllllllllllllll}
\toprule
{} & \multicolumn{3}{c}{FACES $(n,d)=(400,4096)$} && \multicolumn{3}{c}{Colon Tumor $(n,d)=(62,2000)$} && \multicolumn{3}{c}{ALL-AML $(n,d)=(72,7129)$} && \multicolumn{3}{c}{Central Nervous System $(n,d)=(60,7129)$} \\ 
\cline{2-4} \cline{6-8} \cline{10-12} \cline{14-16}
$d'$ & \makecell[c]{\texttt{AgFlow}\\(10000\\Models)} & \makecell[c]{Ridge\_S\\(100\\Models)} & \makecell[c]{Ridge\_C\\(100\\Models)} && \makecell[c]{\texttt{AgFlow}\\(10000\\Models)} & \makecell[c]{Ridge\_S\\(100\\Models)} & \makecell[c]{Ridge\_C\\(100\\Models)} && \makecell[c]{\texttt{AgFlow}\\(10000\\Models)} & \makecell[c]{Ridge\_S\\(100\\Models)} & \makecell[c]{Ridge\_C\\(100\\Models)} && \makecell[c]{\texttt{AgFlow}\\(10000\\Models)} & \makecell[c]{Ridge\_S\\(100\\Models)} & \makecell[c]{Ridge\_C\\(100\\Models)} \\ 
\midrule
  6 & 1135 & 2346 & 2601 && 100 & 281 & 316 && 167 & 8713 & 8841 && 165 & 8531 & 9136 \\ 
  8 & 1284 & 3011 & 3348 && 106 & 361 & 403 && 194 & 11645 & 11722 && 186 & 11350 & 12114 \\ 
  9 & 1369 & 3341 & 3714 && 110 & 400 & 443 && 212 & 13099 & 13158 && 203 & 12758 & 13602 \\ 
  10 & 1440 & 3670 & 4067 && 114 & 439 & 483 && 227 & 14550 & 14601 && 220 & 14167 & 15053 \\ 
  12 & 1598 & 4332 & 4781 && 124 & 526 & 562 && 267 & 17468 & 17475 && 262 & 16992 & 17973 \\ 
  15 & 1830 & 5316 & 5800 && 138 & 637 & 682 && 335 & 21920 & 21776 && 330 & 21217 & 22287 \\ 
  16 & 1915 & 5646 & 6141 && 143 & 676 & 722 && 361 & 23428 & 23209 && 354 & 22628 & 23716 \\ 
  18 & 2086 & 6303 & 6827 && 155 & 754 & 801 && 417 & 26455 & 26074 && 407 & 25446 & 26648 \\ 
  20 & 2246 & 6956 & 7501 && 167 & 833 & 880 && 478 & 29463 & 28931 && 465 & 28280 & 29510 \\ 
  24 & 2600 & 8275 & 8854 && 195 & 991 & 1039 && 610 & 35516 & 34661 && 597 & 34090 & 35210 \\ 
  25 & 2679 & 8597 & 9190 && 204 & 1030 & 1079 && 647 & 37041 & 36093 && 634 & 35554 & 36641 \\ 
  30 & 3127 & 10243 & 10866 && 246 & 1226 & 1276 && 851 & 44606 & 43268 && 836 & 42891 & 43788 \\ 
\bottomrule
\end{tabular}
\end{table}

\subsection{Comparisons of Time Consumption and Performance Tuning}
Table~\ref{table:timeused} illustrates the time consumption of the \texttt{AgFlow} algorithm and Ridge-based algorithms over varying penalties on the four datasets. We can see from the table that the time used in the \texttt{AgFlow} algorithm is only a small portion of that of the Ridge\_S and Ridge\_C which are two versions of Ridge-based algorithms. When the sample size and the number of predictors are both small, as in the Colon Tumor dataset with $(n,d) =(62, 2000)$, the time consumption is acceptable for both \texttt{AgFlow} and Ridge-based algorithms. However, when the number of the dimension becomes extremely large as in the ALL-AML dataset with $(n,d) =(72, 7129)$ or the Central Nervous System data with $(n,d) =(60, 7129)$, the time consumption of Ridge\_S and Ridge\_C becomes dramatically large. For example, when $d'=30$ for the ALL-AML dataset, Ridge\_S requires more than $12.39$ hours, which is unacceptable in practice application, whereas the the \texttt{AgFlow} algorithm requires $14$ minutes, which has dramatically reduced the computation time.  

More specifically, when considering the time consumption of \texttt{AgFlow} and Ridge-Path for the above performance tuning procedure, we can find \texttt{AgFlow} is much more efficient. Table~\ref{table:timeused} shows that \texttt{AgFlow} only consumes $246$ seconds to obtain the estimates of 10,000 penalized PCA models when $d'=30$ for the Colon Tumor data with $d=2000$ genes and $851$ seconds for the ALL-AML Leukemia data with $d=7129$ genes, while Ridge-Path needs $1226$ seconds/$1276$ seconds to obtain only 100 penalized PCA models for the Colon Tumor data and $44,606$ seconds/$43,268$ seconds for the ALL-AML Leukemia data, whether using closed-form Ridge estimators or solver-based ones.

\begin{figure*}
\centering
\includegraphics[width=0.52\textwidth]{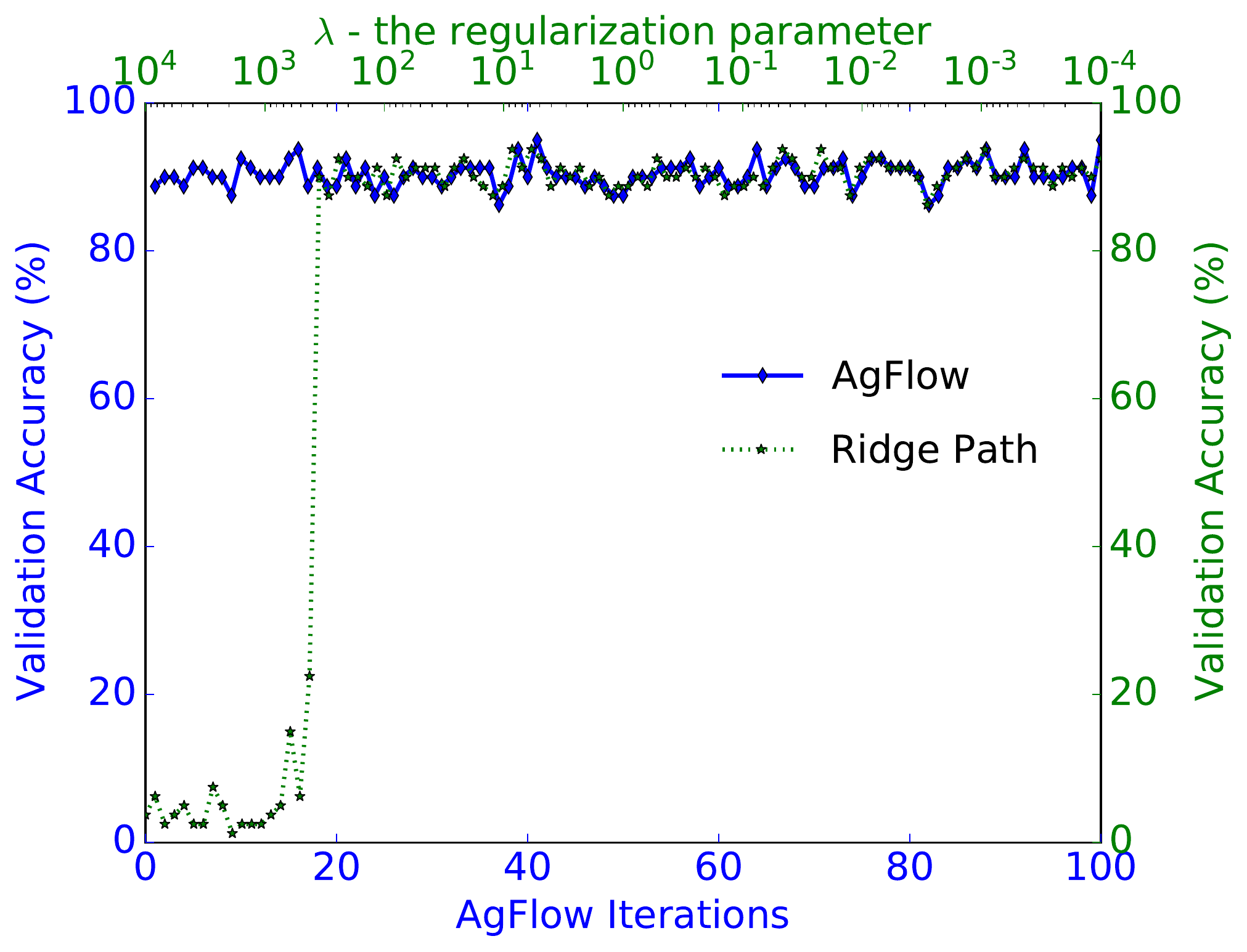}
\caption{An Example of Parameter Tuning on FACES dataset with Random Forest.}
\label{fig:tuning}
\end{figure*}

Fig.~\ref{fig:tuning} illustrates the examples of performance tuning using Ridge-Path and \texttt{AgFlow} over varying penalties with Random Forest classifiers. 
While \texttt{AgFlow} estimates the $\ell_2$-penalized PCA with varying penalty by stopping the SGD optimizer with different number of iterations, Ridge-Path needs to shrink the sample covariance matrix with varying $\lambda$ and estimate $\ell_2$-penalized PCA through the time-consuming matrix inverse. It is obvious that both \texttt{AgFlow} and Ridge-Path have certain capacity to screen models with different penalties.

In conclusion, \texttt{AgFlow} demonstrates both efficiency and effectiveness in model selection for penalized PCA, in comparisons with a wide range of classic and newly-fashioned algorithms~\citep{zou2006sparse,shamir2015stochastic,oja1985stochastic,gene1996matrix}. Note that, the classification accuracy of some tasks here might not be as good as those reported in~\cite{zhu2007markov}. While our goal is to compare the performance of $\ell_2$-penalized PCA model selection with classification accuracy as the selection objective, the work~\cite{zhu2007markov} focus on selecting a discriminative set of features for classification.

\section{Conclusions}\label{sec:conclus}


Since PCA has been widely used for data processing, feature extraction and dimension reduction in unsupervised data analysis, 
we have proposed \texttt{AgFlow} algorithm to do fast model selection with a much lower complexity in $\ell_2$-penalized PCA where the regularization is usually incorporated to deal with the multicolinearity and singularity issues encountered under HDLSS settings. 
Experiments show that our \texttt{AgFlow} algorithm beats the existing methods with an overwhelming improvement with respect to the accuracy and computational complexity, especially, when compared with the ridge-based estimator which is implemented as a time-consuming model estimation and selection procedure among a wide range of penalties with matrix inverse.
Meanwhile, the proposed \texttt{AgFlow} algorithm naturally retrieves the complete solution path of each principal component, which shows an implicit regularization and can help us do the model estimation and selection simultaneously.
Thus we can identify the best model from an end-to-end optimization procedure using low computational complexity. 
In addition, except for the advantage of the accuracy and computational complexity, the \texttt{AgFlow} enlarges the capacities of performance tuning in a more intuitive and easily way. The observations backup our claims. 

\section{Future Work}\label{sec:fut}

Though the \texttt{AgFlow} algorithm naturally retrieves the complete solution path of each principal component and can do model selection under the implicit $\ell_2$-norm regularization effect, the linear combination of all the original variables is often not friendly to interpret the results. New methods with implicit or explicit $\ell_1$-norm regularization (lasso penalty) are in great demand, where $\ell_1$-norm regularization produces sparse solutions and we can do variable estimation and selection simultaneously.

In addition to the Approximated Gradient Flow, we are also interested in the implicit regularization introduced by other (stochastic) optimizers, such as Adam and/or Nesterov's momentum methods, with potential new applications to Markov Chain Monte Carlo or other statistical computations. Furthermore, the implicit regularization of the \texttt{AgFlow} running nonlinear models for statistical inference would be interesting too.

\bibliographystyle{unsrt}
\bibliography{references}   

\end{document}